# Deep-learning based discovery of partial differential equations in integral form from sparse and noisy data

Hao Xu[a], Dongxiao Zhang[b,c,d,*], and Nanzhe Wang[a]

[a] *BIC-ESAT, ERE, and SKLTCS, College of Engineering, Peking University, Beijing 100871, P. R. China*

[b] *Guangdong Provincial Key Laboratory of Soil and Groundwater Pollution Control, School of Environmental Science and Engineering, Southern University of Science and Technology, Shenzhen 518055, P. R. China*

[c] *State Environmental Protection Key Laboratory of Integrated Surface Water-Groundwater Pollution Control, School of Environmental Science and Engineering, Southern, University of Science and Technology, Shenzhen 518055, P. R. China*

[d] *Intelligent Energy Lab, Peng Cheng Laboratory, Shenzhen 518000, P. R. China*

[*] Corresponding author.

E-mail address: 390260267@pku.edu.cn (H. Xu); zhangdx@sustech.edu.cn (D. Zhang); wangnz@pku.edu.cn (N. Wang).

**Abstract**: Data-driven discovery of partial differential equations (PDEs) has attracted increasing attention in recent years. Although significant progress has been made, certain unresolved issues remain. For example, for PDEs with high-order derivatives, the performance of existing methods is unsatisfactory, especially when the data are sparse and noisy. It is also difficult to discover heterogeneous parametric PDEs where heterogeneous parameters are embedded in the partial differential operators. In this work, a new framework combining deep-learning and integral form is proposed to handle the above-mentioned problems simultaneously, and improve the accuracy and stability of PDE discovery. In the framework, a deep neural network is firstly trained with observation data to generate meta-data and calculate derivatives. Then, a unified integral form is defined, and the genetic algorithm is employed to discover the best structure. Finally, the value of parameters is calculated, and whether the parameters are constants or variables is identified. Numerical experiments proved that our proposed algorithm is more robust to noise and more accurate compared with existing methods due to the utilization of integral form. Our proposed algorithm is also able to discover PDEs with high-order derivatives or heterogeneous parameters accurately with sparse and noisy data.

1. Introduction

In the past, models of physical processes, such as the wave equation, the diffusion equation and Burgers equation, are derived from physical laws or summarized from experiments. However, benefited from the rapid development of computer technology, data-driven discovery of partial differential equations (PDEs) has attracted increasing attention, which aims to discover underlying governing equations of physical processes directly from observation data. Without the need of formula derivation, PDE discovery methods can extract possible governing equations directly from data even under complex conditions, which is more effective and time-saving.

In recent investigations, two mainstream methods have been employed to discover unknown PDEs, i.e., sparse regression-based methods and neural network-based methods. For sparse regression based-methods, sparse regression techniques, including least absolute shrinkage and selection operator (LASSO), sequential threshold ridge regression (STRidge) and sparse Bayesian regression, are utilized to identify PDE terms from a pre-determined candidate library that contains several potential terms[1–3]. On the basis of these techniques, numerous works have developed methods in order to deal with broader issues, such as identifying stochastic dynamic systems[4], discovering governing equations for fluid dynamics[5–7], and identifying parametric PDEs[8]. However, although sparse-regression methods are both fast and applicable, they require sufficient data and low noise. For neural network-based methods, neural networks are introduced to solve inverse problems[9–13] or calculate derivatives by automatic differentiation [14–17]. For example, Raissi et al.[18,19] proposed the physics-informed neural network (PINN) to solve the forward and inverse problem, and discover PDEs. Xu et al.[15] combined neural network and genetic algorithm to discover underlying PDEs. Compared with sparse regression-based methods, neural network-based methods are more flexible, accurate, and stable to noise for high accuracy of derivatives calculation via automatic differentiation. Nevertheless, certain shortcomings remain. For example, designing the structure of the neural network is intricate and training a neural network is time-consuming. In addition to these two mainstream methods, Gaussian process[20] and genetic algorithm[21,22] are also utilized to discover unknown PDEs.

Although the above-mentioned approaches are suitable for most cases, there are still some limitations. For example, it is challenging to discover some complex physical dynamics with high order-derivatives (e.g., Kuramoto-Sivashinsky equation with $4^{th}$ derivative) because the calculation of high-order derivatives is inaccurate even with automatic differentiation when data are noisy. It is also difficult for those methods to handle some physical processes in heterogeneous media, where heterogeneous parameters are embedded in the partial differential operators (e.g., the term $\frac{\partial}{\partial x}(K(x)h_x(x,t))$ in the governing equation of flow in heterogeneous media).

Therefore, some recent works have sought to discover the integral form (or weak form) of the PDE instead of the differential form because integration techniques assist to reduce the effect of noise, especially for high-order derivatives[23]. For instance, Somacal et al.[24] discovered differential equations with latent variables by integrating retrieved equations after sparse regression. Reinbold and Grigoriev[25] combined local polynomial interpolation and sparse regression to identify PDEs with latent variables. In their works, however, a large amount of data is required to be placed on spatio-temporal grids for integration. Moreover, although integration techniques are utilized, their methods are not robust to high levels of noise (e.g., 15% noise) because derivatives are calculated by numerical differentiation. In addition, a test function is employed in their works for integration and elimination of the effect of latent variables. However, the test function is selected according to the derivative order, which requires prior information about the target PDE. Although Huang et al.[26] discovered variational laws hidden in physical systems by integral form, the results lack parsimony and redundant terms still exist. Moreover, all of the above-mentioned algorithms do not solve the problem of physical processes in heterogeneous media.

For solving these issues, in this work, an innovative framework is proposed to discover PDEs from sparse and noisy data combining deep-learning and integral form. In the proposed framework, a neural network is employed to approximate sparse observation data, and derivatives are calculated by automatic differentiation. Then, a large number of meta-data are generated by the trained neural network, and derivatives are calculated. Benefited from meta-data, the value of the integral form of PDE terms at any spatial-temporal point can be calculated easily. Finally, the genetic algorithm is employed to discover PDEs in integral form. Numerical experiments are conducted for proof-and-concept, and results prove that our proposed algorithm works well with sparse and noisy data, and is more accurate compared with existing methods. Our proposed method is also suitable for PDEs with high-order derivatives and heterogeneous parameters, and satisfactory outcomes are obtained.

The remainder of this paper proceeds as follows. Section 2 describes the problem statement and the methodology of the proposed framework in detail. Numerical experiments and results are provided in Section 3. Discussions and conclusions are given in Section 4.

## 2. Methodology

*2.1 The differential form of PDE*

In this work, we will consider more kinds of PDEs, including those with high-order derivatives or with heterogeneous parameters, and thus the general differential form of PDE can be described as follows:

$$u_T = \sum_{n=0}^{N} c_n f_n(u, u_x, u_{xx}, u_{xxx}, u_{xxxx}, ...), \qquad (1)$$

where $u_T$ denotes temporal derivatives with various orders, such as $u_t$ or $u_{tt}$; $N$ is the number of terms; $c_n$ is the coefficient of each term, which can be a constant for PDEs with constant coefficients or a variable for PDEs with heterogeneous parameters; and $f_n$ denotes each term in the PDE, which is a multiplication of spatial derivatives of various orders. In previous works[1–3,5,14,21], the differential form of PDE is discovered with the PDE terms $f_n$ and corresponding coefficients $c_n$ identified.

*2.2 Deep neural network approximation*

In this work, a deep neural network (DNN) $NN(x, t; \theta)$ is constructed to approximate the observation data. A DNN usually comprises an input layer, several hidden layers, and an output layer. The forward formulation of DNN can be expressed as:

$$\begin{aligned}
\vec{z}_1 &= \sigma_1(\vec{w}_1 \vec{X} + \vec{b}_1) \\
\vec{z}_2 &= \sigma_2(\vec{w}_2 \vec{z}_1 + \vec{b}_2) \\
&\vdots \\
\vec{z}_l &= \sigma_l(\vec{w}_l \vec{z}_{l-1} + \vec{b}_l) \\
\vec{Y} &= \vec{w}_{l+1} \vec{z}_l + \vec{b}_{l+1}
\end{aligned} \qquad (2)$$

where $l$ is the number of hidden layers; $\vec{X}$ is the input vector; $\vec{Y}$ is the output vector; $\sigma_i$ is the activation function, such as tanh($x$), sin($x$), and Rectified Linear Unit (ReLU); $\vec{w}_i, \vec{b}_i (i=1,2,...,l+1)$ are weights and bias of the neural network; and $\theta = \{\vec{w}_i, \vec{b}_i\}_{i=1}^{l+1}$. The loss function in this work is the mean squared error loss, which can be written as:

$$L(\theta) = \frac{\sum_{j=1}^{N_t} \sum_{i=1}^{N_x} [u(x_i, t_j) - NN(x_i, t_j; \theta)]^2}{N_x N_t}, \qquad (3)$$

where $N_x$ and $N_t$ are the number of $x$ and $t$, respectively. The neural network $NN(x, t; \theta)$ is trained by minimizing the loss function $L(\theta)$ via the optimizer Adam. When the training process has finished, the trained neural network can be utilized to generate meta-data to supplement sparse observation data and calculate derivatives by automatic differentiation.

*2.3 The integral form of PDE*

In this work, we attempt to discover the integral form of PDE. In this part, we will discuss two circumstances, including PDEs with constant coefficients and PDEs with heterogeneous parameters, and a unified integral form that applies to both circumstances is defined.

*2.3.1 Integral form for PDEs with constant coefficients*

Firstly, we consider a simple situation in which the coefficients of target PDEs are constant. Let us integrate Eq. (1) with respect to $x$ for both sides simultaneously. $c_n$ is a constant here, and thus we have the following:

$$\begin{aligned}
\int u_T dx &= \int \sum_{n=0}^{N} c_n f_n(u, u_x, u_{xx}, u_{xxx}, u_{xxxx}, ...) dx \\
&= \sum_{n=0}^{N} c_n \int f_n(u, u_x, u_{xx}, u_{xxx}, u_{xxxx}, ...) dx
\end{aligned} \qquad (4)$$

Then, the integral intervals are referred as $\Omega_k = [x_k - \frac{1}{2}L, x_k + \frac{1}{2}L](k=1,2,...,M)$. Here, $x_k$ denotes the midpoint of each integral interval; $L$ is the length of the integral interval; and $M$ is the number of integral intervals. Therefore, for each integral interval, Eq. (4) can be rewritten to be:

$$\int_{\Omega_k} u_T dx = \sum_{n=0}^{N} c_n \int_{\Omega_k} f_n(u, u_x, u_{xx}, u_{xxx}, u_{xxxx}, ...) dx \tag{5}$$

In this work, the value of integration is calculated by the Gauss-Legendre quadrature formula, which is expressed as:

$$\int_a^b f(x)dx = \frac{b-a}{2} \sum_{l=1}^{N_l} A_l f(\frac{b-a}{2} x_l + \frac{a+b}{2}), \tag{6}$$

where $N_l$, $x_l$, and $A_l$ are the number of integral nodes, the integral nodes, and corresponding coefficients, respectively; and $a$ and $b$ are the lower and upper bounds, respectively. For the Gauss-Legendre quadrature formula, once $N_l$ is determined, $x_l$ and $A_l$ can be fixed. Therefore, the left-hand side of Eq. (5) can be transformed into:

$$\begin{aligned}\int_{\Omega_k} u_T dx &= \int_{x_k - \frac{L}{2}}^{x_k + \frac{L}{2}} u_T dx \\ &= \frac{L}{2} \sum_{l=1}^{N_l} A_l u_T (\frac{L}{2} x_l + x_k, t)\end{aligned} \tag{7}$$

Since the value of $u_T$ at any point can be calculated by generating meta-data and conducting automatic differentiation by the neural network, the value of $\int_{\Omega_k} u_T dx$ for each integral interval can be obtained easily. It is worth noting that meta-data greatly facilitate the integration process, especially with sparse observation data, since they can provide an approximate value at any point, which may not be included in observation data (e.g., the integral nodes of the Gauss-Legendre quadrature formula).

For most governing equations of physical processes, it is easy to convert them from differential form into integral form (or weak form)[27], for example:

$$\begin{gathered}u_t + uu_x + u_{xxx} = 0 \\ \Updownarrow \\ \int_{\Omega_k} u_t dx = \int_{\Omega_k} (-uu_x - u_{xxx}) dx. \\ \Updownarrow \\ \int_{\Omega_k} u_t dx = -\frac{1}{2} u^2 \Big|_{\Omega_k} - u_{xx} \Big|_{\Omega_k}\end{gathered} \tag{8}$$

where the symbol $|_{\Omega_k}$ means the value of the integral term in the integral interval $\Omega_k$, which can also be written as $|_{x_k - \frac{1}{2}L}^{x_k + \frac{1}{2}L}$ in this work. Consequently, PDEs with an explicit integral form are considered in this work, and Eq. (5) can be transformed to be:

$$\int_{\Omega_k} u_T dx = \sum_{n=0}^{N} d_n f'_n \big|_{\Omega_k}, \tag{9}$$

where $f'_n$ is the term of PDE in the integral form, which is also composed of multiplication of spatial derivatives of various orders; and $d_n$ is corresponding coefficients of $f'_n$. For the right-hand side of the integral form, values of $f'_n$ can be easily obtained by calculating the values of these terms at the integral boundary via meta-data and automatic differentiation. It is worth noting that the derivative order required to be calculated will be lower in the integral form, which will assist to improve the accuracy of discovered PDEs.

*2.3.2 Integral form for PDEs with heterogeneous parameters*

In previous works, it is difficult to discover heterogeneous parametric PDEs because heterogeneous parameters are embedded in the partial differential operators. However, in this work, the problem can be solved by integral forms. For most heterogeneous parametric PDEs, it can be written as:

$$u_T = \frac{\partial}{\partial x}\left(\sum_{n=0}^{N} c_n(x) g_n(u, u_x, u_{xx}, u_{xxx}, u_{xxxx}, ...)\right). \tag{10}$$

Different from Eq. (1), $c_n(x)$ is a heterogeneous parameter embedded in the partial differential operator, which may be a random field or a function related to *x*. Similarly, Eq. (10) is integrated with respect to *x* in the integral interval $\Omega_k$ for both sides simultaneously, and its integral form can be expressed as follows:

$$\int_{\Omega_k} u_T dx = \sum_{n=0}^{N} \big[c_n(x) g_n\big]_{\Omega_k}. \tag{11}$$

The only difference between Eq. (11) and Eq. (9) is that the coefficient in Eq. (11) is a variable. Therefore, irrespective of whether the PDEs have constant coefficients or heterogeneous parameters, the integral form is similar, which can be summarized to be a unified form, which is written as:

$$\int_{\Omega_k} u_T dx = \sum_{n=0}^{N} \big(C_n F_n\big)_{\Omega_k}, \tag{12}$$

where $F_n$ denotes the term in the integral form, which is multiplication of spatial derivatives of various orders; and $C_n$ is the corresponding coefficient, which can be a constant or a variable. For PDE discovery in integral form, a parsimonious model is expected to be obtained with the term $F_n$ and corresponding coefficient $C_n$.

With the unified integral form, our proposed method is able to handle both circumstances. After transforming PDEs into integral form, the value of both sides can be calculated, which means that the integral form of PDE can be also discovered by commonly-used PDE discovery methods, such as sparse regression or genetic algorithm.

*2.4 Genetic algorithm*

The genetic algorithm is a widely used optimization method in the field of machine learning and adaptive control[22,28]. Compared with sparse regression methods, the genetic algorithm is able to discover PDEs with an incomplete initial library via mutation and cross-over that can produce infinite combinations of genes[16]. Therefore, the genetic algorithm is employed to discover the PDE in integral form. In this part, we will briefly introduce the procedure of the genetic algorithm, including digitization, fitness calculation, mutation, cross-over, and evolution. For additional details, please refer to Xu et al.[15].

*2.4.1 Digitization*

In the genetic algorithm, the process of digitization aims to construct mapping between genes and corresponding terms. The basic component in the genetic algorithm is the gene, which is represented numerically. In this work, the number refers to the order of derivative. For example, for the right-hand side, we have the following:

$$0 \leftrightarrow u,\ 1 \leftrightarrow u_x,\ 2 \leftrightarrow u_{xx}.$$

Similarly, for the left-hand side, we also have:

$$1 \leftrightarrow \int u_t dx,\ 2 \leftrightarrow \int u_{tt} dx.$$

After the definition of the gene, the gene module is then constructed by combining genes via multiplication. For example:

$$[1,3] \leftrightarrow u_x u_{xxx},\ [0,1] \leftrightarrow uu_x.$$

Finally, the genome can be generated by the combination of gene module via addition. For each genome, there is a unique PDE in integral form corresponding to it. Here, the gene module is expressed by braces for the right-hand side in order to distinguish the left and right sides of the PDE. For example:

$$\int u_t dx = u^2 + u_x \leftrightarrow [1], \{[0,0],[1]\}.$$

Through the principle of digitization, we can randomly generate genomes and corresponding PDEs with several basic genes.

*2.4.2 Fitness calculation*

In the genetic algorithm, fitness is the measurement the quality of the genome. In this work, the fitness function is defined as follows:

$$Fitness = MSE + \varepsilon \cdot L_{genome}, \tag{13}$$

where $L_{genome}$ denotes the length of the genome; and $\varepsilon$ is a hyper-parameter, which is selected according to the magnitude of *MSE*. For each genome, the structure of the corresponding PDE can be translated through the principle of digitization. Meanwhile, values of each term can be obtained by the trained neural network. Therefore, the coefficients of terms can be calculated by least squared regression, and *MSE* is the mean squared error, which is defined as:

$$MSE = \frac{\left\| \vec{U}_T - \vec{U}_X \vec{\beta} \right\|^2}{N_{meta}}, \tag{14}$$

with

$$\vec{U}_T = \begin{bmatrix} U_T(x_1,t_1) \\ U_T(x_2,t_1) \\ \vdots \\ U_T(x_{N_x^{meta}},t_1) \\ U_T(x_1,t_2) \\ \vdots \\ U_T(x_{N_x^{meta}},t_{N_t^{meta}}) \end{bmatrix}, \vec{U}_X = \begin{bmatrix} U_X(x_1,t_1) \\ U_X(x_2,t_1) \\ \vdots \\ U_X(x_{N_x^{meta}},t_1) \\ U_X(x_1,t_2) \\ \vdots \\ U_X(x_{N_x^{meta}},t_{N_t^{meta}}) \end{bmatrix}. \tag{15}$$

where $\vec{\beta}$ is the vector of coefficients calculated by the least squared regression; $N_x^{meta}$ and $N_t^{meta}$ are the number of spatial and temporal observation points in the meta-data that is generated on a spatial-temporal grid, respectively; and $N_{meta}$ is the number of meta-data, and we have:

$$N_{meta} = N_x^{meta} \cdot N_t^{meta}.$$

$U_T$ and $U_X$ are the matrix of the left-hand and right-hand side terms of the PDE in integral form, respectively, and we have the following:

$$U_T(x_i,t_j) = \int_{x_i-\frac{L}{2}}^{x_i+\frac{L}{2}} u_T(x,t_j)dx; \; U_X(x_i,t_j) = [F_0\Big|_{x_i-\frac{L}{2},t_j}^{x_i+\frac{L}{2},t_j}, F_1\Big|_{x_i-\frac{L}{2},t_j}^{x_i+\frac{L}{2},t_j}, \cdots, F_n\Big|_{x_i-\frac{L}{2},t_j}^{x_i+\frac{L}{2},t_j}].$$

Therefore, the size of $U_T$ is $N_{meta} \times 1$, the size of $U_X$ is $N_{meta} \times n$, and the size of $\vec{\beta}$ is $n \times 1$. With the fitness function, for each genome, its fitness can be calculated according to Eq. (13). In this work, the lower is the fitness, the better is the model.

*2.4.3 Cross-over and mutation*

Cross-over and mutation are crucial for the genetic algorithm because they are able to produce new genomes from parent genomes, which greatly increases the search range of the genetic algorithm. For cross-over, certain gene modules in two different genomes are exchanged. For mutation, there are three ways, including order mutation, add-module mutation and delete-module mutation, in this work. For order mutation, the order of derivatives in the gene will be reduced by 1 and 0 can be mutated to the highest order considered. For add-module mutation, a random module will be added to the genome. For delete-module mutation, a random module will be deleted from the genome. The process of cross-over and mutation will occur under certain probability.

*2.4.4 Evolution*

In the process of evolution, parent genomes cross-over twice to produce a filial generation, the number of which is twice as many as parent genomes. Afterwards, mutation takes place, and the fitness of the genome in the filial generation is calculated. Finally, the first half of the children with the smallest fitness is preserved to be the parent genomes of the subsequent generation. The evolution process will continue until reaching the maximum generations.

*2.5 The stepwise method*

For the unified integral form Eq. (12), the coefficients may be constants or variables. However, most PDE discovery methods can only handle PDEs with constant coefficients. Therefore, for PDEs with heterogeneous parameters, a stepwise method is employed, which aims to discover the structure of the PDE first, and then adjust the coefficients. Here, the heterogeneous parameters are assumed to be related to $x$ because parameters are heterogeneous for space in most cases. The diagrammatic sketch of our proposed algorithm for discovering PDEs with heterogeneous parameters is shown in Fig. 1.

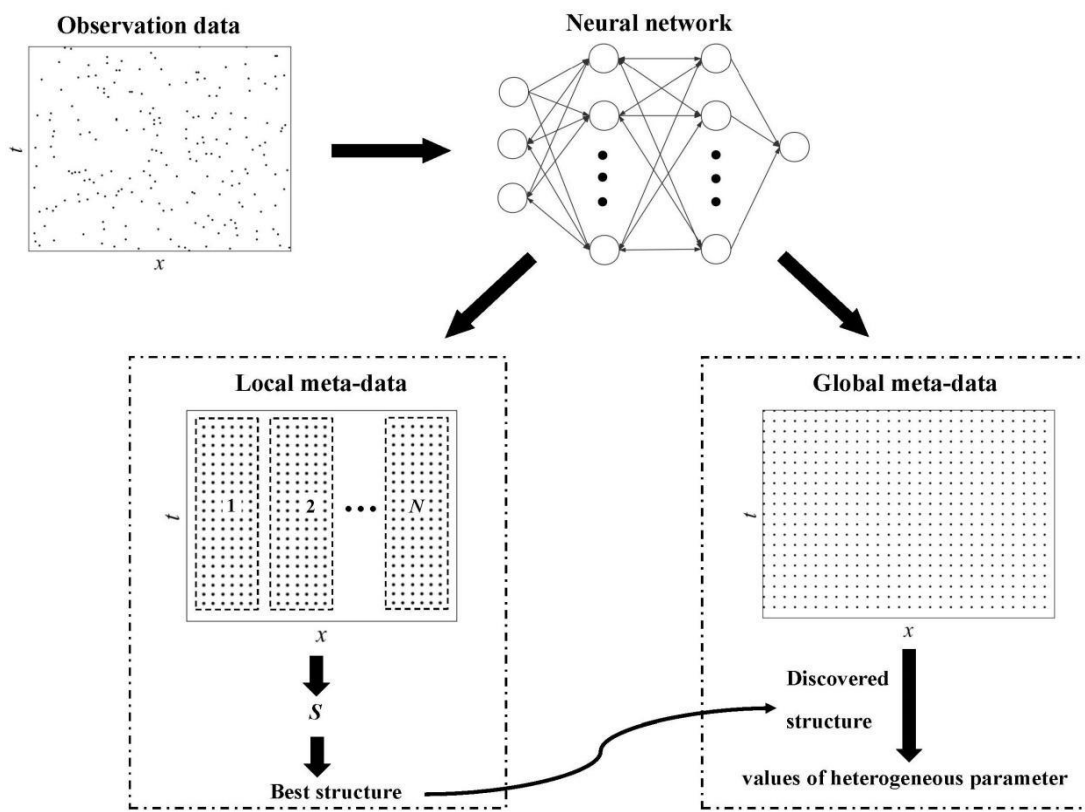

**Fig. 1.** Diagram of the proposed algorithm, with black dots referring to data. The local meta-data are generated on the local window, the global meta-data are generated on nearly the entire domain, and both of them are on a spatial-temporal grid. The best structure is obtained from discovered structures in each local window. The value of heterogeneous parameters is calculated based on the best structure and global meta-data.

*2.5.1 Discovery of PDE structure*

In the stepwise method, the entire spatial domain is uniformly divided into several subdomains, called local windows. For each local window, local meta-data are generated, and the genetic algorithm is utilized to discover the structure of the PDE. Among these structures, the structure that occurs most frequently is termed the best (or most possible) structure. Here, a stability function is defined as:

$$S = \frac{N_{best}}{N_{local}}, \qquad (16)$$

where $S$ is the stability; $N_{best}$ is the count of the occurrence of the best structure; and $N_{local}$ is the number of local windows. A larger $S$ indicates higher confidence of the discovered PDE structure. Additional details about the stepwise method can be found in Xu et al.[16].

*2.5.2 Calculating heterogeneous parameters*

Although the structure of PDE has been discovered, the coefficients are imprecise, and should be adjusted based on the discovered structure and global meta-data generated in a larger domain. Since it is unknown whether each coefficient of the discovered term is a constant or a variable, it is assumed that all coefficients are varying at first. Here, the discovered structure is written as follows:

$$\int_{\Omega_k} u_T dx = \sum_{n=0}^{N_{terms}} [C_n(x) F_n]_{\Omega_k}, \qquad (17)$$

where $F_n$ and $N_{terms}$ are the term and the number of terms in the discovered structure, respectively. For a one-dimensional problem, $\Omega_k = [x_k - \frac{1}{2}L, x_k + \frac{1}{2}L]$, and thus Eq. (17) can be rewritten as:

$$\int_{\Omega_k} u_T dx = \sum_{n=0}^{N_{terms}} \left( C_n(x_k + \frac{1}{2}L) F_n(x_k + \frac{1}{2}L, t) - C_n(x_k - \frac{1}{2}L) F_n(x_k - \frac{1}{2}L, t) \right). \qquad (18)$$

In order to calculate the value of $C_n(x)$ on each point, the integral interval $L$ is fixed to be $2\Delta x$ with $\Delta x = x_{k+1} - x_k$. Since meta-data are generated on a uniformly distributed spatial-temporal grid, $\Delta x$ is a constant. With this particular value of $L$, Eq. (18) can be transformed into multi-element coupled linear equations:

$$\int_{\Omega_k} u_T(x_k, t_j)dx = \sum_{n=0}^{N_{terms}} \left( C_n(x_{k+1})F_n(x_{k+1}, t_j) - C_n(x_{k-1})F_n(x_{k-1}, t_j) \right), \tag{19}$$

with $k=2,3,...,N_x^{meta}-1$ and $j=1,2,...,N_t^{meta}$. In multi-element coupled linear equations, $C_n(x_k)$ is unknown, and thus the count of unknown number is $N_{terms}(N_x^{meta}-2)$. Meanwhile, the number of equations is $N_t^{meta}(N_x^{meta}-2)$. Therefore, Eq. (19) are overdetermined equations when $N_t^{meta} > N_{terms}$. In this work, $N_t^{meta}$ is much larger than possible $N_{terms}$, which means that Eq. (19) are overdetermined equations, and the optimal solution can be easily obtained by the least square method.

After the values of the coefficient for each term at each spatial point have been calculated, a criterion is defined to differentiate between constant and heterogeneous parameters. The coefficient of variation is defined as:

$$cv = \left| \frac{\sigma_{coef}}{\mu} \right| \times 100\% \tag{20}$$

where $\sigma_{coef}$ is the standard deviation of calculated coefficients; and $\mu$ is the mean of coefficients. If $cv$<5%, the coefficient will be identified as a constant, and its value equals to $\mu$.

## 3. Results

In this subsection, our approach will be examined by several numerical experiments that highlight different aspects of the problem. Firstly, we will assume that the coefficients of the PDE are known to be constants in order to better assess the performance of our approach when handling high-order derivatives with noisy and discrete data. Finally, discovery of PDEs with heterogeneous parameters with our approach is investigated. It is worth noting that whether or not the parameter of the underlying physical process is heterogeneous is unknown here, which is closer to the actual situation.

In this work, a five-layer ANN with 50 neurons per hidden layer is utilized to fit the observation data, and the activation function is sin(*x*). The Gauss-Legendre quadrature formula with five integral nodes is used for integration. For the genetic algorithm, the population size of genomes is 200, and the number of maximum generations is 100. Basic genes are set to be $[\int u_t dx(1), \int u_{tt} dx(2)]\{u(0), u_x(1), u_{xx}(2), u_{xxx}(3)\}$. The probability of cross-over and mutation is 0.8 and 0.2, respectively. The entire dataset in this work is generated by numerical simulation, which is detailed in Appendix A.

*3.1 Discovery of PDEs with high-order derivatives using discrete and noisy data*

In this part, the Korteweg-de Vries (KdV) equation with 3rd derivative and the KS equation with 4th derivative are considered to examine the performance of our proposed algorithm. The hyper-parameter $\varepsilon$ is set to be $10^{-3}$.

*3.1.1 Discovery of the KdV equation*

The KdV equation is utilized to describe one-way motion of shallow water, which is written as:

$$u_t = -uu_x - 0.0025u_{xxx}. \tag{21}$$

For the dataset, there are 512 spatial observation points in the domain $x \in [-1,1)$, and 201 temporal observation points in the domain $t \in [0,1]$. Therefore, the total number of data is 102,912. For meta-data, there are 1,000 spatial points in the domain $x \in [-0.5, 0.5)$, and 200 temporal observation points in the domain $t \in [0,1]$, and thus the total number of meta-data is 200,000. In this experiment, the integral interval $L$ is 0.05, and different volumes of data are randomly selected to train the neural network. Meanwhile, the error of discovered PDE is defined as:

$$error = \frac{\sum_{j=1}^{N_t}\sum_{i=1}^{N_x}\left|u(x_i,t_j) - u'(x_i,t_j)\right|^2}{N_x N_t}, \tag{22}$$

where $u(x_i,t_j)$ is the solution of correct PDE; $u'(x_i,t_j)$ is the solution of discovered PDE; and $N_x$ and $N_t$ are the number of $x$ and $t$, respectively. The result is shown in Table 1.

Table 1. The KdV equation identified in integral form with different volumes of data training the neural network.

| Correct PDE: | $u_t = -uu_x - 0.0025u_{xxx}$ $\Updownarrow$ $\int u_t dx = -0.5u^2 - 0.0025u_{xx}$ | |
|---|---|---|
| **Data Volume** | **Learned Equation** | **Error** |
| 30,000 (29.15%) datapoints | $\int u_t dx = -0.497u^2 - 0.00249u_{xx}$ | 2.92% |
| 15,000 (14.58%) datapoints | $\int u_t dx = -0.499u^2 - 0.00249u_{xx}$ | 1.30% |
| 5,000 (4.859%) datapoints | $\int u_t dx = -0.497u^2 - 0.00248u_{xx}$ | 3.51% |
| 1,000 (0.972%) datapoints | $\int u_t dx = -0.496u^2 - 0.00248u_{xx}$ | 4.08% |
| 500 (0.486%) datapoints | $\int u_t dx = -0.490u^2 - 0.00244u_{xx}$ | 10.99% |

| 100 (0.097%) datapoints | $\int u_t dx = -0.340u^2 - 0.00158u_{xx} + 3.76 \times 10^{-7} u_x u_{xxx}$ | -- |

From the table, it can be seen that our proposed algorithm is able to discover the correct PDE, even with 500 datapoints which only account for 0.486% of the total datapoints. In addition, the error is relatively small. In previous works[1,14,15], although the KdV equation can be identified, the amount of data required is large due to the existence of the 3rd derivative. With the integral form, the order of the derivative in the equation can be reduced, which will simultaneously decrease the demand for data and maintain high accuracy of the result.

*3.1.2 Discovery of the KS equation*

The KS equation is a partial differential equation which reflects the comprehensive influence of gas diffusion and heat conduction on the stability of the plane flame front. Its form is written as follows:

$$u_t = -uu_x - u_{xx} - u_{xxxx}. \tag{23}$$

For the dataset, there are 512 spatial observation points in $x \in [-10,10)$, and 251 temporal observation points in $t \in [0,50]$. Consequently, the total number of data is 128,512. For meta-data, there are 1,200 spatial points in $x \in [-6,6]$ and 300 temporal observation points in $t \in [10,40]$, and thus the total number of meta-data is 360,000. In this experiment, the integral interval $L$ is 1. Firstly, 60,000 data are randomly selected to train the neural network, and different levels of noise are added to the data in the following form:

$$u(x,t) = u(x,t) \cdot (1 + \gamma \times e), \tag{24}$$

where $\gamma$ denotes the noise level; and $e$ is the uniform random variable, taking values from -1 to 1. The result is shown in Table 2.

Table 2. The KS equation identified in integral form with different levels of noise added to the data.

| Correct PDE: | $u_t = -uu_x - u_{xx} - u_{xxxx}$ $\Updownarrow$ $\int u_t dx = -0.5u^2 - u_x - u_{xxx}$ | |
|---|---|---|
| **Noise Level** | **Learned Equation** | **Error** |
| Clean data | $\int u_t dx = -0.480u^2 - 0.947u_x - 0.951u_{xxx}$ | 25.12% |
| 1% noise | $\int u_t dx = -0.464u^2 - 0.907u_x - 0.913u_{xxx}$ | 41.32% |
| 5% noise | $\int u_t dx = -0.355u^2 - 0.633u_x - 0.656u_{xxx}$ | 94.87% |
| 10% noise | $\int u_t dx = 0.131 u_x$ | -- |

From the table, it is found that our proposed method is robust to 5% noise. However, the error is relatively large because, even though the 4th order derivative is reduced to 3rd order through the integral form, its order remains high and is difficult to calculate precisely by automatic differentiation with discrete data. In contrast, it is worth noting that in previous works which aim to discover the differential form[1,3], they are only robust to 0.5% noise due to huge inherent errors resultant from calculating 4th derivatives with limited data.

Next, the performance of our proposed method when discovering the KS equation with different volumes of data is investigated. 60,000 (46.7%) datapoints, 24,000 (18.7%) datapoints, 12,000 (9.34%) datapoints and 6,000 (4.67%) datapoints are randomly selected to train the neural network, and the integral form is discovered by our proposed method, respectively. The result is shown in Table 3.

Table 3. The KS equation identified in integral form with different volumes of data.

| Correct PDE: | $u_t = -uu_x - u_{xx} - u_{xxxx}$ $\Updownarrow$ $\int u_t dx = -0.5u^2 - u_x - u_{xxx}$ | |
|---|---|---|
| Data Volume | Learned Equation | Error |
| 60,000 (46.69%) datapoints | $\int u_t dx = -0.480u^2 - 0.947u_x - 0.951u_{xxx}$ | 25.12% |
| 24,000 (18.68%) datapoints | $\int u_t dx = -0.475u^2 - 0.935u_x - 0.940u_{xxx}$ | 33.69% |
| 12,000 (9.337%) datapoints | $\int u_t dx = -0.475u^2 - 0.937u_x - 0.941u_{xxx}$ | 31.27% |
| 6,000 (4.669%) datapoints | $\int u_{tt} dx = 0.082 uu_x u_{xx} + 0.088 u^2 u_x + 0.108 u_x$ | -- |

From the table, it can be seen that our proposed method is able to discover the correct PDE with relatively high accuracy, even with 12,000 (9.34%) discrete datapoints. Meanwhile, the performance of the method is stable until the volume of data is too small to calculate the 3rd derivative accurately. It is worth noting that in previous works[1,3], although the KS equation can also be discovered with relatively high accuracy, a huge amount of data is required to be distributed on a regular spatial-temporal grid.

3.2 Comparison with the differential form

In this part, the integral form is compared with the differential form in terms of discovering PDEs with high-order derivatives. The KdV equation is considered again. 30,000 data are randomly selected to train the neural network. For the integral form, our proposed method is utilized to discover the PDE. For the differential form, the genetic algorithm is directly employed to discover PDEs without the integration process. In order to better demonstrate the distinction of the results of both methods, different levels of noise are added to the data. The outcomes are displayed in

Table 4.

Table 4. The KdV equation identified in integral form and in differential form with different levels of noise added to data, respectively.

| Correct PDE: | $u_t = -uu_x - 0.0025u_{xxx}$ $\Updownarrow$ $\int u_t dx = -0.5u^2 - 0.0025u_{xx}$ | | | |
|---|---|---|---|---|
| **Noise Level** | **Learned Equation in Integral Form** | **Error** | **Learned Equation in Differential Form** | **Error** |
| Clean data | $\int u_t dx = -0.497u^2 - 0.00249u_{xx}$ | 2.92% | $u_t = -0.990uu_x - 0.00247u_{xxx}$ | 5.50% |
| 1% noise | $\int u_t dx = -0.494u^2 - 0.00247u_{xx}$ | 6.12% | $u_t = -0.987uu_x - 0.00247u_{xxx}$ | 6.49% |
| 5% noise | $\int u_t dx = -0.493u^2 - 0.00247u_{xx}$ | 6.90% | $u_t = -0.977uu_x - 0.00244u_{xxx}$ | 11.87% |
| 10% noise | $\int u_t dx = -0.481u^2 - 0.00239u_{xx}$ | 20.37% | $u_t = -0.937uu_x - 0.00232u_{xxx}$ | 32.99% |
| 15% noise | $\int u_t dx = -0.472u^2 - 0.00234u_{xx}$ | 29.49% | $u_t = -0.788uu_x - 0.00191u_{xxx}$ | 86.45% |
| 20% noise | $\int u_t dx = -0.370u^2 - 0.00178u_{xx}$ | 95.50% | $u_t = 0.215u^2 u_x - 0.7uu_x - 0.0019u_{xxx} + 0.0006uu_{xxx}$ | -- |
| 25% noise | $\int u_t dx = -0.346u^2 - 0.00162u_{xx}$ | 101.99% | $u_t = -3.4 \times 10^{-4} uu_x u_{xx} - 0.48uu_x - 0.0017u_{xxx} + 5.4 \times 10^{-4} uu_{xxx}$ | -- |

From the table, the distinction between the integral form and the differential form can be seen clearly. Firstly, discovering PDEs in the integral form is more robust to noise (robust to 25% noise) compared with the differential form (only robust to 15% noise), which means that the process of integration is able to diminish the influence of data noise to a certain extent. In addition, faced with the same data and the same level of noise, the discovered PDEs in the integral form through our proposed method is more accurate. This is because the order of derivatives needed to be calculated in the PDE is decreased in the integral form, which will reduce the error of the calculation of derivatives when data are noisy.

*3.3 Discovery of PDEs with heterogeneous parameters*

In actual cases, the underlying physical process usually occurs in heterogeneous media, where heterogeneous parameters are embedded in the partial differential operators. In this work, a more difficult situation is considered, in which the heterogeneous parameter is a random field following a specific distribution with corresponding covariance $\sigma_R^2$. The Karhunen–Loeve expansion (KLE) is introduced here to parameterize the heterogeneous parameter, and the finite difference method is employed to generate the dataset. Additional details about the random field and KLE are provided in Appendix B. It is worth noting that, our proposed method can also handle other simple situation where the heterogeneous parameter is a function related to $x$ similarly. In the numerical experiments below, the hyper-parameter $\varepsilon$ is set to be $10^{-5}$, the number of local window is 10, and the integral interval $L = 2\Delta x$.

*3.3.1 The heterogeneous parametric convection-diffusion equation*

Firstly, let us consider the heterogeneous parametric convection-diffusion equation, which can be utilized to describe the contaminant transport process in heterogeneous media. Its form is written as follows:

$$u_t = \frac{\partial}{\partial x}(Du_x + vu), \qquad (25)$$

where $D$ is a random field with the variance $\sigma_R^2 = 1$; 12 eigenvectors are considered in KLE to approximate the random field; $v$ is a constant; and $v$=-1. For the dataset, there are 801 spatial observation points in $x \in [0,8]$, and 251 temporal observation points in $t \in [0,2.5]$. Therefore, the total number of data is 201,051. 100,000 data are randomly selected to train the neural network. For local meta-data, there are 200 observation points in $x \in [0.8n, 0.8(n+1)](n = 0,1,...9)$ and 100 observation points in $t \in [0.25, 2.25]$, and thus the number of meta-data is 20,000 for each local window. For global meta-data, there are 400 observation points in $x \in [1,7]$, and 300 observation points in $t \in [0.25, 2.25]$. Consequently, the number of global meta-data is 120,000. The structure discovered in each local window is shown in Table 5.

Table 5. The structure of PDE discovered in each local window for discovering the heterogeneous parametric convection-diffusion equation.

| Local Window | Discovered Structure |
|---|---|
| $x \in [0, 0.8]$ | $\int u_t dx, u^2$ |
| $x \in [0.8, 1.6]$ | $\int u_t dx, u, u^2$ |
| $x \in [1.6, 2.4]$ | $\int u_t dx, u, u_x$ |
| $x \in [2.4, 3.2]$ | $\int u_t dx, u, u_x$ |
| $x \in [3.2, 4.0]$ | $\int u_t dx, u, u_x$ |
| $x \in [4.0, 4.8]$ | $\int u_t dx, u, u_x$ |

| $x \in [4.8, 5.6]$ | $\int u_t dx, u, u_x$ |
|---|---|
| $x \in [5.6, 6.4]$ | $\int u_t dx, u, u_x$ |
| $x \in [6.4, 7.2]$ | $\int u_t dx, u, u_x$ |
| $x \in [7.2, 8.0]$ | $\int u_t dx, u, u_x$ |

From the table, it is clear that the best structure is $\int u_t dx, u, u_x$ and $S = 0.8$, which indicates that the discovered structure has a high degree of confidence. With the best structure, the value of heterogeneous parameter can be calculated by solving Eq. (19). The result is shown in Fig. 2.

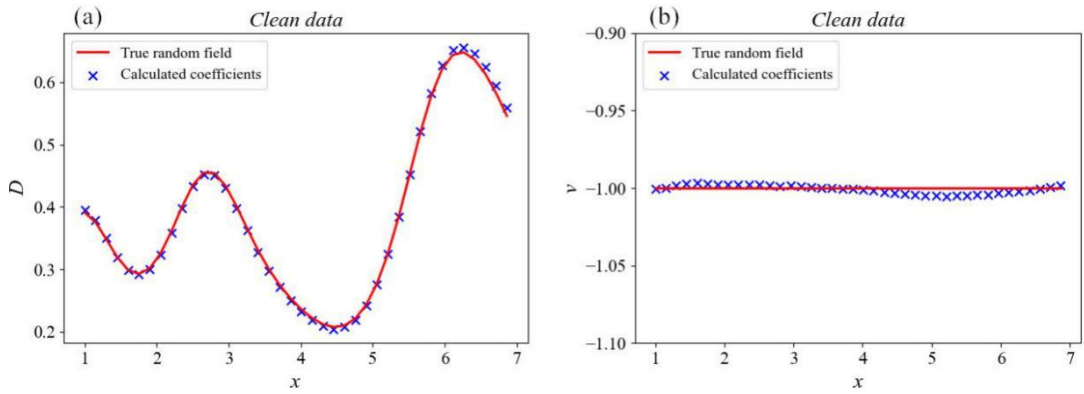

**Fig. 2**. The scatter of calculated parameters $D$ (a) and $v$ (b), and corresponding curve of the true values for the heterogeneous parametric convection-diffusion equation, respectively.

For the two parameters, the standard deviation of the parameter $D$ ($\sigma_D$) is 0.1389, the mean of the parameter $D$ ($\mu_D$) is 0.3894, and the coefficient of variation ($cv_D$) is calculated to be 35.67%. Moreover, $\sigma_v$ is 0.0027, $\mu_v$ is -1.0008, and $cv_v$ is calculated to be 0.268%. Therefore, $v$ is identified to be a constant for $cv_v < 5\%$, and its value equals to $\mu_v$.

From the above, it can be seen that the structure of PDE is identified correctly, and the values of heterogeneous parameters are calculated accurately. Meanwhile, the constant parameter is successfully distinguished by our proposed algorithm. It is worth noting that this kind of PDE is hard to be discovered using existing methods because the heterogeneous parameters are embedded in the partial differential operators, and thus the differential form will lead to many more terms with varying coefficients, which is not concise.

*3.3.2 The heterogeneous parametric wave equation*

Next, the random parametric wave equation is considered, which is usually utilized to describe the vibration of heterogeneous elastic rods. Its form is expressed as follows:

$$u_{tt} = \frac{\partial}{\partial x}(EA(x)u_x), \quad (26)$$

where *EA* refers to tensile stiffness and is a random field with the variance $\sigma_R^2 = 1$. 12 eigenvectors are considered in KLE to approximate the random field, and no noise is added. For the dataset, there are 401 spatial observation points in $x \in [0,8]$, and 251 temporal observation points in $t \in [0,6]$. Consequently, the total number of data is 100,651. 90,000 data are randomly selected to train the neural network. For local meta-data, there are 200 observation points in $x \in [0.8n, 0.8(n+1)] (n = 0,1,...9)$ and 100 observation points in $t \in [0.5, 5.5]$, and thus the number of meta-data is 20,000 for each local window. For global meta-data, there are 400 observation points in $x \in [1,7]$, and 300 observation points in $t \in [0.5, 5.5]$. Therefore, the number of global meta-data is 120,000. The structure discovered in each local window is shown in Table 6.

Table 6. The structure of PDE discovered in each local window for the heterogeneous parametric wave equation.

| Local Window | Discovered Structure |
|---|---|
| $x \in [0, 0.8]$ | $\int u_{tt} dx, u_x$ |
| $x \in [0.8, 1.6]$ | $\int u_{tt} dx, u_x$ |
| $x \in [1.6, 2.4]$ | $\int u_{tt} dx, u_x$ |
| $x \in [2.4, 3.2]$ | $\int u_{tt} dx, u_x$ |
| $x \in [3.2, 4.0]$ | $\int u_{tt} dx, u_x$ |
| $x \in [4.0, 4.8]$ | $\int u_{tt} dx, u_x$ |
| $x \in [4.8, 5.6]$ | $\int u_{tt} dx, u_x$ |
| $x \in [5.6, 6.4]$ | $\int u_{tt} dx, u_x$ |
| $x \in [6.4, 7.2]$ | $\int u_{tt} dx, u_x$ |
| $x \in [7.2, 8.0]$ | $\int u_{tt} dx, u_x$ |

From the table, it is clear that the best structure is $\int u_{tt}dx, u_x$ and $S=1$. With the best structure, the value of the parameter can be calculated by solving Eq. (19). The result is shown in Fig. 3.

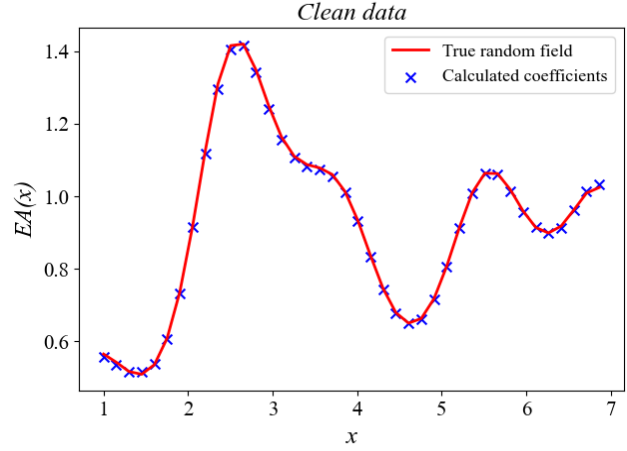

**Fig. 3**. The scatter of the calculated parameter and the curve of the true values for the heterogeneous parametric wave equation.

Meanwhile, $\sigma_{EA}$ is 0.2403, $\mu_{EA}$ is 0.92960, and $cv_{EA}$ is calculated to be 25.85%. Consequently, $v$ is identified to be a heterogeneous parameter for $cv_{EA} > 5\%$. From the above, it is found that the structure of PDE is discovered stably, and the calculated coefficients have high accuracy. Moreover, the left-hand side of the heterogeneous parametric wave equation is $u_{tt}$ and is also discovered, which means that our proposed method can handle different kinds of left-hand side terms.

*3.3.3 Discovery of heterogeneous parametric PDEs with noisy data*

In this part, discovery of heterogeneous parametric PDEs with noisy data is investigated, and the heterogeneous convection-diffusion equation is considered again. The conditions are the same as those in Section 3.3.1. Different levels of noise, including clean data, 1% noise, 5% noise, 10% noise and 15% noise, are added to noise, and our proposed method is employed to discover the structure and calculate the corresponding heterogeneous parameters. With different levels of noise, the discovered best structure, the stability $S$, the variance and mean of calculated parameters, and the coefficient of variation are displayed in Table 7, respectively. In addition, the value of calculated parameters $D$ and $v$ are shown in Fig. 4 and Fig. 5, respectively.

**Table 7**. The discovered best structure, the stability, the variance and mean of calculated parameters, and the coefficient of variation for discovery of the heterogeneous parametric convection-diffusion equation with different levels of noise.

| Noise Level | Clean Data | 1% Noise | 5% Noise | 10% Noise | 15% Noise |
|---|---|---|---|---|---|
| Discovered Best Structure | $\int u_t dx, u_x, u$ | $\int u_t dx, u_x, u$ | $\int u_t dx, u_x, u$ | $\int u_t dx, u_x, u$ | $\int u_t dx, u_x, u$ |

| $s$ | 0.8 | 0.8 | 0.8 | 0.8 | 0.8 |
|---|---|---|---|---|---|
| $\sigma_D$ | 0.1389 | 0.1369 | 0.1364 | 0.1438 | 0.1451 |
| $\mu_D$ | 0.3894 | 0.3880 | 0.3865 | 0.3861 | 0.3862 |
| $cv_D$ | 35.67% | 35.28% | 35.29% | 37.24% | 37.58% |
| $\sigma_v$ | 0.0027 | 0.0026 | 0.0051 | 0.0133 | 0.0260 |
| $\mu_v$ | -1.0008 | -0.9996 | -0.9993 | -1.0027 | -0.9955 |
| $cv_v$ | 0.268% | 0.260% | 0.510% | 1.326% | 2.613% |

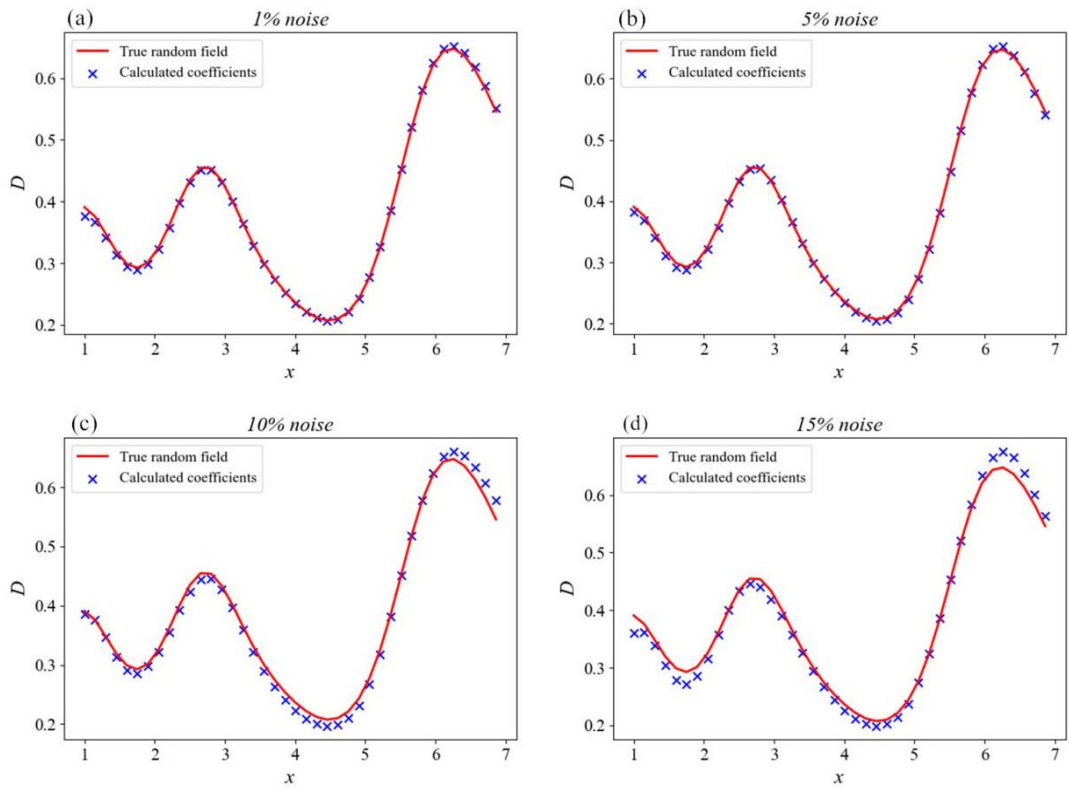

**Fig. 4**. The scatter of calculated heterogeneous parameters $D$ and corresponding curve of the true values for the heterogeneous parametric convection-diffusion equation with (a) 1% noise, (b) 5% noise, (c) 10% noise, and (d) 15% noise, respectively.

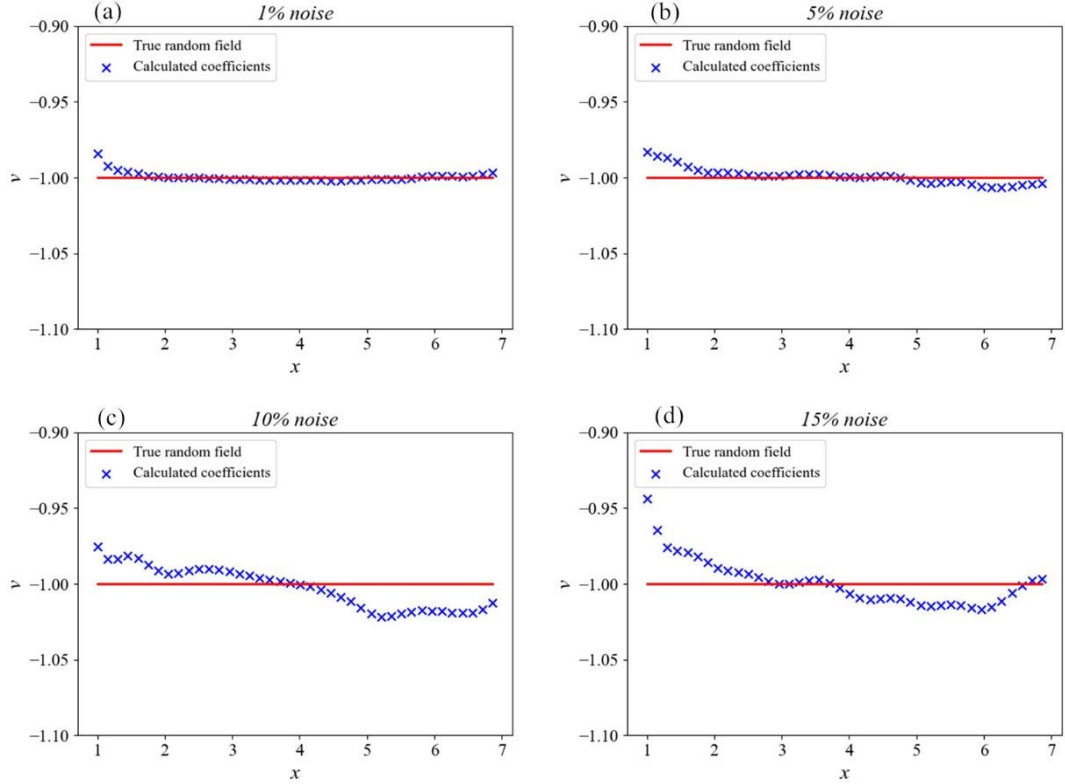

**Fig. 5**. The scatter of calculated parameter *v* and corresponding curve of the true values for the heterogeneous parametric convection-diffusion equation with (a) 1% noise, (b) 5% noise, (c) 10% noise, and (d) 15% noise, respectively.

From the table and the figures, it can be found that our proposed method is extremely robust to noise when discovering heterogeneous parametric PDEs. Even with 15% noise, the best structure is discovered stably, and the value of parameter *v* which is identified to be a constant is accurate. Furthermore, although the value of heterogeneous parameter *D* has some deviations when the noise is high, the error is not large and is acceptable.

*3.3.4 Discovery of heterogeneous parametric PDEs with sparse data*

In this part, the performance of our proposed method for discovering heterogeneous parametric PDEs with sparse data is investigated. The heterogeneous parametric wave equation is considered again. Different from Section 3.3.2, different volumes of data, including 25,000 (24.8%) datapoints, 5,000 (4.97%) datapoints, 1,000 (0.99%) datapoints and 500 (0.50%) datapoints, are randomly selected to train the neural network. Our proposed method is employed to discover the best structure and calculate the parameter. For all cases, the results show that the discovered best structure is $\int u_{tt}dx, u_x$ with the stability $S = 1$, which means that our proposed algorithm is able to discover the correct structure stably with extremely sparse data. The scatter of the calculated parameter is shown in Fig. 6.

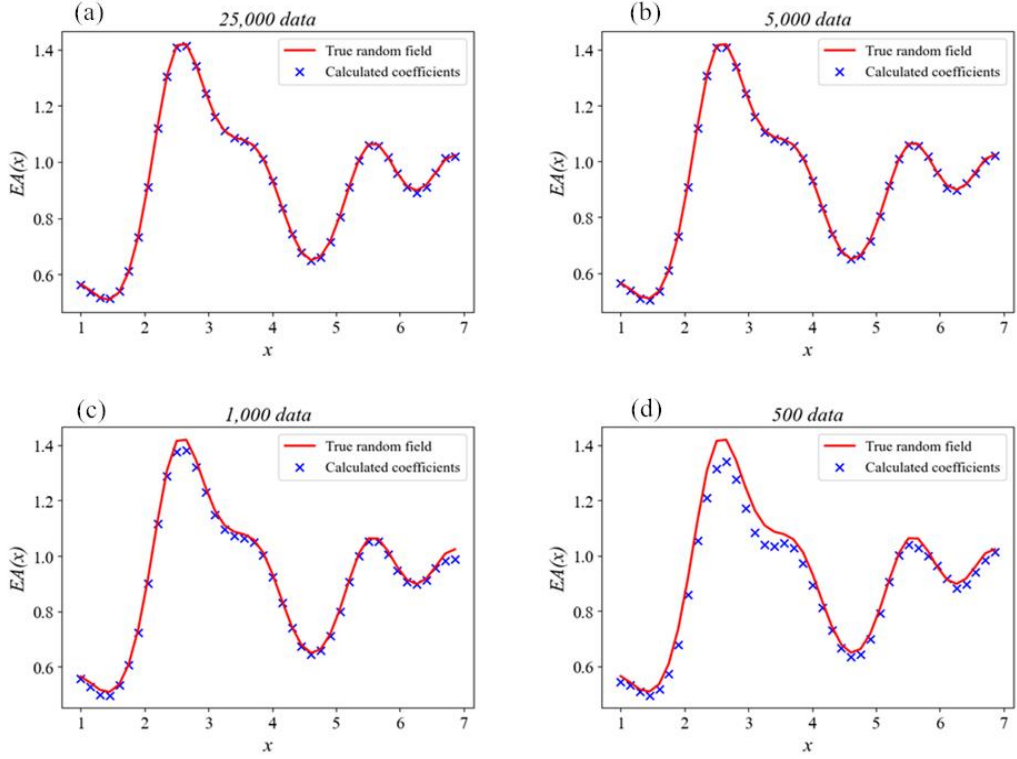

**Fig. 6**. The scatter of the calculated parameter and the curve of true values for the heterogeneous parametric wave equation with different volumes of data, including 25,000 datapoints (a), 5,000 datapoints (b), 1,000 datapoints (c), and 500 datapoints (d) training the neural network, respectively.

From Fig. 6, it can be seen that with only 5,000 (4.97%) datapoints, the values of heterogeneous parameter can be calculated accurately. When the available datapoints are extremely sparse (e.g., 1,000 datapoints), although the values of heterogeneous parameter may not be highly accurate, the structure is stably discovered. It is worth noting that, for PDEs with varying parameters that are more complex, a large amount of data are required in previous works[8,16]. Therefore, it is proven that our proposed method can handle sparse data even when discovering heterogeneous parametric PDEs.

*3.3.5 The influence of the amplitude of the heterogeneous parameter*

In this part, the performance of our proposed method for discovering heterogeneous parametric PDEs with higher-amplitude parameters is investigated. The heterogeneous parametric wave equation is considered again, and the settings are the same as those in Section 3.3.2. The amplitude of the heterogeneous parameter is tuned by changing the variation of the random field ($\sigma_R^2$) when generating the random field. The dataset where the random field is generated with $\sigma_R^2$=0.5, 1, 2 and 4 is utilized to train the neural network, and our proposed method is employed to discover the best structure and corresponding parameters, respectively. With different magnitudes of the heterogeneous parameter, the discovered best structure, the stability $S$, the variance and mean of calculated parameters, and the coefficient of variation are displayed in Table 8, respectively. Meanwhile, the value of calculated parameters is shown in Fig. 7. Results demonstrate that our proposed method is able to discover the best structure correctly and stably when the magnitude is higher. Moreover, the values of calculated parameters are accurate even with high magnitude.

**Table 8**. The discovered best structure, the stability, the variance and mean of calculated parameters, and the coefficient of variation for discovery of the heterogeneous parametric wave equation with different magnitude of heterogeneous parameter.

| $\sigma_R^2$ | **0.5** | **1** | **2** | **4** |
|---|---|---|---|---|
| **Discovered Best Structure** | $\int u_{tt}dx, u_x$ | $\int u_{tt}dx, u_x$ | $\int u_{tt}dx, u_x$ | $\int u_{tt}dx, u_x$ |
| $s$ | 1 | 1 | 1 | 1 |
| $\sigma_{EA}$ | 0.1757 | 0.2043 | 0.3244 | 0.4363 |
| $\mu_{EA}$ | 0.9436 | 0.9296 | 0.9133 | 0.9062 |
| $cv_{EA}$ | 18.62% | 25.85% | 35.52% | 48.15% |

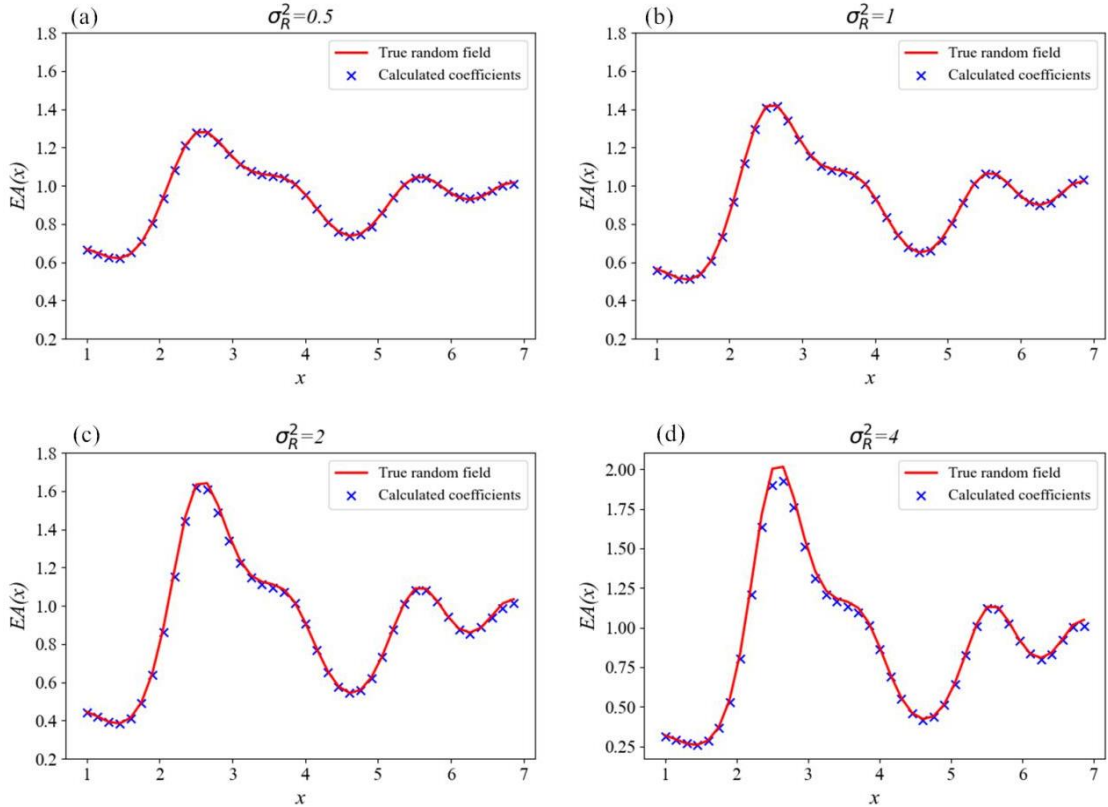

**Fig. 7**. The scatter of the calculated parameter and the curve of true values for the heterogeneous parametric wave equation with different magnitudes of heterogeneous parameter. The picture contains the calculated parameter $EA$ with (a) $\sigma_R^2 = 0.5$, (b) $\sigma_R^2 = 1$, (c) $\sigma_R^2 = 2$, and (d) $\sigma_R^2 = 4$, respectively.

## 4. Conclusion and Discussion

In this work, we proposed a novel framework which combines deep-learning and the integral form to discover the governing equation of underlying PDEs. Instead of discovering the differential form as in most previous works, we intended to identify a unified integral form of PDE which is suitable for PDEs with both constant and varying coefficients. In our proposed method, a deep neural network is trained by observation data first. Next, the trained neural network is utilized to generated meta-data and calculate derivatives via automatic differentiation, and then values of terms in integral form are calculated. Finally, the genetic algorithm is employed to discover the best structure and the value of corresponding coefficients. The numerical experiments proved that our proposed method is able to handle various types of PDEs, including PDEs with high derivatives and heterogeneous parameters, with sparse and noisy data.

Compared with previous works that discover the differential form, the order of the derivative needed to be calculated in the integral form is lower, which increases the accuracy of discovered PDEs because the calculation of derivative is more accurate when the order of derivative is low. As a consequence, satisfactory performance of our proposed method is obtained for identifying PDEs with high-order derivatives, such as the KdV equation and the KS equation. Numerical experiments also demonstrate that our proposed method is more robust to noise. There are three main reasons that account for the satisfactory performance with high noise. Firstly, the derivatives are calculated by automatic differentiation of the neural network, which can smooth the noise to a certain extent. Secondly, the process of integration is able to reduce the effect of noise. Finally, the integral form decreases the order of the derivative in the target PDE, and the lower order derivatives are less affected by noise. As a consequence, our proposed method is able to discover the correct PDE with 5% noise for the KS equation and 25% noise for the KdV equation.

Different from other works concerning the integral form, the value of integration is calculated on meta-data generated on a spatial-temporal grid by the trained neural network rather than on the observation data, which means that our proposed algorithm is especially suitable for dealing with limited and sparse data. The numerical experiment proved this assertion, since our proposed method successfully discovered the KdV equation with only 500 (0.486%) discrete datapoints. In contrast, a large number of data are needed to calculate the 3$^{rd}$ derivative in previous works. The influence of integral interval length is also investigated in Appendix C, and the result shows that our proposed method is stable to discover the correct PDE with a wide range of integral interval length.

In addition, our proposed method can identify PDEs with heterogeneous parameters, which are always utilized to describe physical processes in heterogeneous media. For this kind of PDE, the heterogeneous parametric parameters are always embedded in the partial operator, which is difficult to be identified in differential form. This is because the PDE will have many terms and become a non-concise form after expansion, and thus no research has yet been conducted on this problem. In this work, the integral form is employed to solve this problem by extracting the heterogeneous parameters after integration. A harder situation is considered in which the heterogeneous parameters are random fields, and the numerical experiment demonstrated that our proposed method performs well even with sparse data, noisy data, and relatively large amplitudes of heterogeneous parameters. It is worth noting that our proposed method is able to identify whether the parameter is constant or heterogeneous, as well, which means that the method can work with little prior knowledge. Moreover, for heterogeneous parametric PDEs with stronger nonlinearity, our proposed algorithm can also discover the correct structure and calculate the value of heterogeneous parameters accurately, although the stability may be relatively low, which is detailed in Appendix D.

Despite the advantages, our proposed algorithm still possesses certain limitations and challenges. Firstly, for some governing equations of complex dynamics, their corresponding integral forms are unable to be represented by the combination of different orders of spatial derivatives. Therefore, it is difficult for our proposed method to discover this kind of PDE. Secondly, our proposed method is time-consuming, since it needs to integrate on each integral interval. Finally, the choice of hyper-parameter will affect the discovered PDE, which is decided by experiments and is adjusted by comparing the posterior error, which involves solving PDEs and is sophisticated. Further investigation of these issues is necessary.

**Acknowledgements**

This work is partially funded by the National Natural Science Foundation of China (Grant No. 51520105005 and U1663208) and the National Science and Technology Major Project of China (Grant No. 2017ZX05009-005 and 2017ZX05049-003).

# Appendix A: Additional details about the dataset of the discussed PDEs

## 1. The Korteweg–de Vries (KdV) equation

The KdV equation takes the following form:

$$u_t = -uu_x - 0.0025 u_{xxx}. \tag{A.1}$$

In order to generate the dataset, the conventional spectral method is employed to numerically solve the KdV equation. The initial condition is set to be $u(0,x) = \cos(\pi x)$ with $x \in [-1,1]$, and the boundary conditions are assumed to be periodic. We integrate Eq. (A.1) from $t=0$ to $t=1$ utilizing the Chebfun package with a spectral Fourier discretization with 512 modes and a fourth-order explicit Runge-Kutta temporal integrator with time-step size being $10^{-6}$. The solution is recorded every $\Delta t = 0.005$ to obtain 201 observation steps in time. Therefore, we have 512 spatial observation points and 201 temporal observation points, and the total number of the dataset is 102,912.

## 2. The Kuramoto-Sivashinsky (KS) equation

The KS equation takes the form:

$$u_t = -uu_x - u_{xx} - u_{xxxx}. \tag{A.2}$$

In order to generate the dataset, the KS equation is solved utilizing the finite difference method. For the dataset, $x \in [-10,10]$ and $t \in [0,50]$. The initial condition is set to be $u(0,x) = \sin(-\frac{\pi}{10}x)$, and the boundary condition is $u(t,-10) = u(t,10) = 0$. The PDE is solved by the second-order Runge-Kutta method with 1,024 spatial observation points for 250,001 timesteps. Then, temporal observation points are taken every 1,000 timesteps. Therefore, we have 512 spatial observation points and 251 temporal observation points, and the total number of the dataset is 128,512.

## 3. The heterogeneous parametric convection diffusion equation

The heterogeneous parametric convection diffusion equation takes the following form:

$$u_t = \frac{\partial}{\partial x}(Du_x + vu). \tag{A.3}$$

In order to generate the dataset, Eq. (A.3) is solved utilizing the finite difference method. For the dataset, $x \in [0,8]$ and $t \in [0,2.5]$. The initial condition is set to be $u(0,x) = (8-x)\sin(x)$, and the boundary condition is $u(t,0) = u(t,8) = 0$. The PDE is solved with 801 spatial observation points for 250,001 timesteps. Then, temporal observation points are taken every 1,000 timesteps. Therefore, we have 801 spatial observation points and 251 temporal observation points, and the total number of the dataset is 201,051.

**4. The heterogeneous parametric wave equation**

The heterogeneous parametric wave equation takes the form:

$$u_{tt} = \frac{\partial}{\partial x}(EA(x)u_x). \tag{A.4}$$

In order to generate the dataset, Eq. (A.4) is solved utilizing the finite difference method. For the dataset, $x \in [0,8]$ and $t \in [0,6]$. The initial condition is set to be $u(0,x) = \frac{1}{2}\sin(\frac{\pi}{4})$, and the boundary condition is $u(t,0) = u(t,8) = 0$. The PDE is solved with 401 spatial observation points for 250,001 timesteps. Then, temporal observation points are taken every 1,000 timesteps. Therefore, we have 401 spatial observation points and 251 temporal observation points, and the total number of the dataset is 100,651.

**Appendix B: The random field and the Karhunen–Loeve expansion (KLE)**

In this work, several stochastic partial differential equations are discovered in which the heterogeneous parameter is assumed to be a random field. The random field can be parameterized by the KLE. For a random field $R(x,\tau) = \ln K(x,\tau)$ where $x$ belongs to a physical domain $D$ and $\tau$ belongs to a probability space, it can be rewritten as:

$$R(x,\tau) = \overline{R}(x) + \tilde{R}(x,\tau), \tag{B.1}$$

where $\overline{R}(x)$ is the mean of the random field; and $\tilde{R}(x,\tau)$ is the fluctuation. The spatial structure of the random field can be described by the covariance $C_R(x,\tilde{x}) = \langle \tilde{R}(x,\tau)\tilde{R}(\tilde{x},\tau) \rangle$. Since the covariance is bounded, symmetric and positive-definite, it can be decomposed as follows:

$$C_R(x,\tilde{x}) = \sum_{i=1}^{\infty} \lambda_i f_i(x) f_i(\tilde{x}), \tag{B.2}$$

where $\lambda_i$ and $f_i(x)$ are the eigenvalue and eigenfunction, respectively, which is calculated by numerically solving the second-type Fredholm equation:

$$\int_D C_R(x,\tilde{x}) f(x) = \lambda f(\tilde{x}). \tag{B.3}$$

Thus, Eq. (B.1) can be transformed into:

$$R(x,\tau) = \overline{R}(x) + \sum_{i=1}^{\infty} \sqrt{\lambda_i} f_i(x) \xi_i(\tau), \tag{B.4}$$

where $\xi_i(\tau)$ are orthogonal Gaussian random variables with zero mean and unit variance. Although there are infinite terms in Eq. (B.4), we can arrange the eigenvalues from large to small, and take the first $n$ eigenvalues to approximate the random field, and thus we have:

$$R(x,\tau) \approx \overline{R}(x) + \sum_{i=1}^{n}\sqrt{\lambda_i}f_i(x)\xi_i(\tau), \tag{B.5}$$

With a larger $n$, more energy (or information) of the random field will be reserved, and the approximated random field will be closer to the true one.
In this work, we set the covariance function to be:

$$C_R(x,\tilde{x}) = \sigma_R^2 e^{-\frac{|x-\tilde{x}|}{\eta}}, \tag{B.6}$$

where $\sigma_R^2$ and $\eta$ are the variance and the correlation length of the process, respectively. For this covariance function, there is an analytical solution for Eq. (B.3), which is written as:

$$\lambda_i = \frac{2\eta\sigma_R^2}{\eta^2\omega_i^2 + 1} \tag{B.7}$$

and

$$f_i(x) = \frac{1}{\sqrt{(\eta^2\omega_i^2+1)L/2+\eta}}[\eta\omega_i\cos(\omega_i x) + \sin(\omega_i x)], \tag{B.8}$$

where $\omega_i$ are the positive roots of the characteristic equation

$$(\eta^2\omega^2 - 1)\sin(\omega L) = 2\eta\omega\cos(\omega L). \tag{B.9}$$

where $L$ is the length of domain $D$. Through adjusting $\sigma_R^2$, the magnitude of the simulated random field can be changed. In this work, the value of the heterogeneous parameter is generated by KLE. $n$ is set to be 12, and 95.6% energy is reserved. $\eta$ is set to be $0.4L$, and $\sigma_R^2$ is set to be 1 initially and is adjusted to change the magnitude of the simulated random field. $\xi_i(\tau)$ is randomly generated by the function *random* in the python module *numpy*. The random seed is 520 for heterogeneous parametric convection-diffusion equation and 2008 for heterogeneous parametric wave equation.

**Appendix C: The influence of integral interval length**

In Section 3.1, the integral interval $L$ is also a hyper-parameter for discovering the KdV and KS equations. In this part, the influence of the integral interval length is investigated. The KdV equation is considered as an example, and the conditions are the same as those in Section 3.1.1. 30,000 data are randomly selected to train the neural network, and 1% noise are added to the data. Our proposed method is employed to discover the KdV equation with $L$=0.025, 0.05, 0.1, 0.2, and 0.5, respectively. The results are displayed in Table C1.

Table C1. The learned equation and corresponding error for discovering the KdV equation with different integral intervals.

| Correct PDE: | $u_t = -uu_x - 0.0025u_{xxx}$ $\Updownarrow$ $\int u_t dx = -0.5u^2 - 0.0025u_{xx}$ | |
|---|---|---|
| **Integral Interval Length** | **Learned Equation** | **Error** |
| 0.025 | $\int u_t dx = -0.4945u^2 - 0.002473u_{xx}$ | 5.58% |
| 0.05 | $\int u_t dx = -0.4941u^2 - 0.002467u_{xx}$ | 6.12% |
| 0.1 | $\int u_t dx = -0.4984u^2 - 0.002493u_{xx}$ | 1.58% |
| 0.2 | $\int u_t dx = -0.4988u^2 - 0.002495u_{xx}$ | 1.18% |
| 0.5 | $\int u_t dx = -0.4999u^2 - 0.002496u_{xx}$ | 0.76% |

From the table, it can be seen that our proposed method is able to discover the correct PDE with a wide range of integral interval length, which means that the choice of integral interval length is not strict. It is also found that for noisy data, a relatively larger integral interval length will assist to enhance the accuracy of the learned equation.

## Appendix D: Discovery of heterogeneous parametric Boussinesq equation

The above experiments examined the performance of our proposed method when discovering heterogeneous parametric PDEs. In this part, the ability of our proposed method to handle complex PDEs with stronger nonlinearity is examined. The heterogeneous parametric Boussinesq equation is considered. This PDE is always utilized to describe the fluctuation of groundwater in heterogeneous aquifers near the shore. Its form is written as follows:

$$u_t = \frac{\partial}{\partial x}(K(x)uu_x), \tag{D.1}$$

where *K* denotes the permeability, which is a heterogeneous parameter here. This PDE has not yet been discovered for proof-and-concept because it has stronger nonlinearity, and its differential form is very complex.

In order to generate the dataset, Eq. (D.1) is solved utilizing the finite difference method. For the dataset, $x \in [0,1]$ and $t \in [0,1]$. The initial condition is set to be $u(0, x) = x\sin(\pi x)$, and the boundary condition is $u(t, 0) = 3t\sin(\pi t); u(t,1) = 2(1-t)\sin(\pi t)$. The PDE is solved with 401 spatial observation points for 250,001 timesteps. Then, temporal observation points are taken every 1,000 timesteps. Therefore, we have 401 spatial observation points and 251 temporal observation points, and the total number of the dataset is 100,651. For local meta-data, there are 200 observation points in $x \in [0.1, 0.1(n+1)] (n = 0,1,...9)$ and 100 observation points in $t \in [0.02, 0.98]$, and thus the number of meta-data is 20,000 for each local window. For global meta-data, there are 400 observation points in $x \in [0.1, 0.9]$, and 300 observation points in $t \in [0.02, 0.98]$. Therefore, the number of global meta-data is 120,000. The structure discovered in each local window is shown in Table D1.

Table D1. The structure of PDE discovered in each local window for the heterogeneous parametric Boussinesq equation.

| Local Window | Discovered Structure |
|---|---|
| $x \in [0, 0.1]$ | $\int u_t dx, u$ |
| $x \in [0.1, 0.2]$ | $\int u_t dx, u u_x$ |
| $x \in [0.2, 0.3]$ | $\int u_t dx, u^2 u_x$ |
| $x \in [0.3, 0.4]$ | $\int u_t dx, u u_x$ |
| $x \in [0.4, 0.5]$ | $\int u_t dx, u u_x$ |
| $x \in [0.5, 0.6]$ | $\int u_t dx, u^3$ |
| $x \in [0.6, 0.7]$ | $\int u_t dx, u_x$ |
| $x \in [0.7, 0.8]$ | $\int u_t dx, u u_x$ |
| $x \in [0.8, 0.9]$ | $\int u_t dx, u^2 u_x$ |
| $x \in [0.9, 1.0]$ | $\int u_t dx, u^2$ |

.

From the table, it is found that the best structure is $\int u_t dx, u u_x$ and $S = 0.4$. Here, although the best structure is the correct structure, the $S$ is relatively low, which indicates that the stability of the discovered PDE is relatively low. With the best structure, the value of heterogeneous parameters can be calculated by solving Eq. (19). The result is shown in Fig. D1.

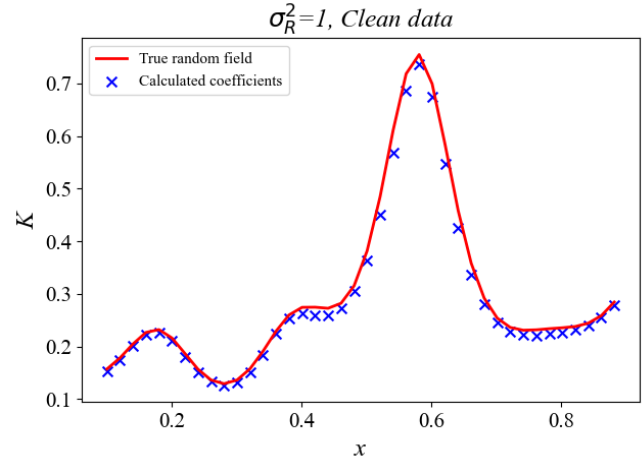

**Fig. D1**. The scatter of calculated parameter and the curve of the true value for the heterogeneous parametric Boussinesq equation.

For the calculated heterogeneous parameters, the standard deviation $\sigma_K$ is 0.1519, the mean $\mu_K$ is 0.2896, and the coefficient of variation $cv_K$ is calculated to be 52.44%. From the above results, although the stability of the discovered best structure is relatively low, the value of heterogeneous parameters remains accurate. This means that our proposed method is also able to handle PDEs with high non-linearity, although the stability may be relatively low when identifying the structure.